\def\year{2018}\relax
\documentclass{article} 
\usepackage{aaai18}  
\usepackage{times}  
\usepackage{helvet}  
\usepackage{courier}  
\usepackage{url}  
\usepackage{graphicx}  
\usepackage[usenames,dvipsnames,svgnames,table]{xcolor}
\usepackage{mdframed}
\usepackage{amssymb}
\usepackage{amsmath}
\usepackage{proof}  
\usepackage{bussproofs}
\usepackage{verbatim}

\usepackage[amsmath,thmmarks]{ntheorem}
\usepackage{relsize}
\usepackage{mathtools}    
\DeclarePairedDelimiterX\setc[2]{\{}{\}}{\,#1 \;\delimsize\vert\; #2\,}

\newcommand{\IGNORE}[1]{}

\newcommand{\lsort}[1]{%
  \ensuremath{\mbox{\textsf{#1}}}}
\newcommand{\defsort}[2]{%
  \newcommand{#1}{\lsort{#2}}}
\defsort{\Action}{Action}
\defsort{\Time}{Time}
\defsort{\Self}{Self}

\defsort{\Agent}{Agent}
\defsort{\Entrant}{Entrant}
\defsort{\ActionType}{ActionType}
\defsort{\Moment}{Moment}
\defsort{\Boolean}{Formula}
\defsort{\PayOut}{PayOut}
\defsort{\Fluent}{Fluent}
\defsort{\Event}{Event}
\defsort{\Object}{Object}
\defsort{\RealTerm}{RealTerm}

\defsort{\Numeric}{Numeric}
\defsort{\Number}{Number}
\defsort{\Trolley}{Trolley}
\defsort{\Track}{Track}
\defsort{\Moveable}{Moveable}
\defsort{\Situation}{Situation}

\newcommand{\lsymbol}[1]{%
  \ensuremath{\mathit{#1}}}
\newcommand{\defsymbol}[2]{%
  \newcommand{#1}{\lsymbol{#2}}}
\defsymbol{\action}{action}
\defsymbol{\initially}{initially}
\defsymbol{\holds}{Holds}
\defsymbol{\happens}{happens}
\defsymbol{\clipped}{clipped}
\defsymbol{\initiates}{initiates}
\defsymbol{\terminates}{terminates}
\defsymbol{\prior}{prior}
\defsymbol{\interval}{interval}

\defsymbol{\does}{does}
\defsymbol{\plans}{plans}
\defsymbol{\act}{act}
\defsymbol{\react}{react}
\defsymbol{\payTot}{pay_{tot}}
\defsymbol{\fight}{fight}
\defsymbol{\coop}{coop}
\defsymbol{\enter}{enter}
\defsymbol{\stayout}{stayout}
\defsymbol{\learns}{learns}
\defsymbol{\payoff}{payoff}
\defsymbol{\position}{position}
\defsymbol{\dead}{dead}
\defsymbol{\damaged}{damaged}

\defsymbol{\onrails}{onrails}
\defsymbol{\switch}{switch}
\defsymbol{\drop}{drop}
\defsymbol{\ins}{in}
\defsymbol{\actionsit}{Action}
\defsymbol{\sick}{sick}
\defsymbol{\visit}{visit}

\newcommand{\lconstant}[1]{%
  \ensuremath{\mbox{\textsf{#1}}}}
\newcommand{\defconstant}[2]{%
  \newcommand{#1}{\lconstant{#2}}}
\defconstant{\Enter}{Enter}
\defconstant{\StayOut}{StayOut}
\defconstant{\Fight}{Fight}
\defconstant{\Acquiesce}{Acquiesce}
\defconstant{\cs}{cs }

\defconstant{\john}{John}
\defconstant{\doctor}{doctor}

\newcommand{\lmodality}[1]{%
  \ensuremath{\mathbf{#1}}}
\newcommand{\defmodality}[2]{%
  \newcommand{#1}{\lmodality{#2}}}
\defmodality{\common}{C}
\defmodality{\knows}{K}
\defmodality{\believes}{B}
\defmodality{\perceives}{P}
\defmodality{\mental}{M}

\defmodality{\desires}{D}
\defmodality{\intends}{I}
\defmodality{\says}{S}
\defmodality{\ought}{O}

\mathchardef\mhyphen="2D
\newcommand{\english}[1]{\textit{{``#1''}}}

\newcommand{\sep}{\ \lvert \ }
\newcommand{\ConsF}[1]{\ensuremath{\mathsf{Cons}[\{#1\}] }}
\newcommand{\Cons}[1]{\ensuremath{\mathsf{Cons}[#1] }}

\newcommand{\Ia}{\ensuremath{\mathbf{I}_1}}
\newcommand{\Ib}{\ensuremath{\mathbf{I}_2}}
\newcommand{\Ic}{\ensuremath{\mathbf{I}_3}}

\newcommand{\II}{\ensuremath{\mathsf{I}}}

\usepackage{booktabs}
\usepackage{mdframed}
\usepackage{paralist}

\usepackage{color}

\usepackage[usenames,dvipsnames,svgnames,table]{xcolor}

\newcommand{\DDE}{\ensuremath{\mathcal{{DDE}}}}
\newcommand{\DCEC}{\ensuremath{{\mathcal{{DCEC}}}}}

\newcommand{\type}[1]{\textsf{#1}}

\newcommand{\Knows}{\ensuremath{\mathbf{K}}}

\newcommand{\intro}[1]{\ensuremath{#1\mhyphen\mathit{intro}}}
\newcommand{\elim}[1]{\ensuremath{#1\mhyphen\mathit{elim}}}

\newcommand{\EmptySet}{\ensuremath{\{\}}}

\newcommand{\counterfac}[2]{\ensuremath{#1 \hookrightarrow #2}}
\newcommand{\matcond}[2]{\ensuremath{#1 \rightarrow #2}}

\begin{document}
%
\title{Counterfactual Conditionals in Quantified
Modal Logic}
\date{\today}
\author{Naveen Sundar Govindarajulu \normalfont{and}
  \textbf{Selmer Bringsjord}\\ 
Rensselaer Polytechnic Institute, Troy, NY  \\
\{naveensundarg,selmer.bringsjord\}@gmail.com}
\maketitle
\begin{abstract}
  We present a novel formalization of counterfactual conditionals in a quantified modal logic.  Counterfactual conditionals play a vital role in ethical and moral reasoning.  Prior work has shown that moral reasoning systems (and more generally, theory-of-mind reasoning systems) should be at least as expressive as first-order (quantified) modal logic (QML) to be well-behaved.  While existing work on moral reasoning has focused on counterfactual-\emph{free} QML moral reasoning, we present a fully specified and implemented formal system that includes counterfactual conditionals.  We validate our model with two projects.  In the first project, we demonstrate that our system can be used to model a complex moral principle, the doctrine of double effect.  In the second project, we use the system to build a data-set with true and false counterfactuals as licensed by our theory, which we believe can be useful for other researchers.  This project also shows that our model can be computationally feasible.
\end{abstract}

\section{Introduction}

Natural-language counterfactual conditionals (or simply
counterfactuals) are statements that have two parts (semantically, and
sometimes syntactically): an antecedent and a consequent.
Counterfactual conditionals differ from standard material conditionals
in that the mere falsity of the antecedent does not lead to the
conditional being true.  For example, the sentence \english{If John
  had gone to the doctor, John would not be sick now} is considered a
counterfactual as it is usually uttered when \english{John has not
  gone to the doctor}.  Note that the surface syntactic form of such
conditionals might not be an explicit conditional such as \english{If
  X then Y}; for example: \english{John going to the doctor would have
  prevented John being sick now}.  Material conditionals in classical
logic fail when used to model such sentences.  Counterfactuals occur
in certain ethical principles and are often associated with moral
reasoning.  We present a formal computational model for such
conditionals.

The plan for the paper is as follows.  We give a brief overview of how
counterfactuals are used, focusing on moral reasoning.  Then we
briefly discuss prior art in modeling counterfactuals, including
computational studies of counterfactuals.  We then present our formal
system, used as a foundation for building our model of counterfactual
conditionals.  After this, we present the model itself and prove some
general properties of the system.  We end by discussing two projects
to demonstrate how the formal model can be used.

\section{Use of Counterfactual Conditionals}
Counterfactual reasoning plays a vital role in human moral reasoning.
For instance, the \textbf{doctrine of double effect} (\DDE) requires
counterfactual statements in its full formalization.  \DDE\ is an
attractive target for building ethical machines, as numerous empirical
studies have shown that human behavior in moral dilemmas is in
accordance with what the doctrine (or modified versions of it)
predict.\footnote{Moral dilemmas are situations in which all available
  actions have large positive and negative effects.} Another reason
for considering \DDE\ is that many legal systems use the doctrine for
defining criminality.  We briefly state the doctrine below.  Assume
that we have an ethical hierarchy of actions as in the \emph{deontological}
case (e.g.\ forbidden, morally neutral, obligatory); see
\cite{sep_deontic_logic}.\footnote{There are also exist in the
  literature more fine-grained hierarchies\cite{ethical_hierarchy_icre2015}.}
We also assume that we have a utility or
goodness function for states of the world or effects, as in the
\emph{consequentialist} case.  Given an agent $a$, an action $\alpha$ in a
situation $\sigma$ at time $t$ is said to be $\DDE$-compliant
\emph{iff} (the clauses are verbatim from \cite{nsg_sb_dde_2017}):

\begin{small}

\begin{mdframed}[linecolor=white, frametitle= \DDE\  Informal, frametitlebackgroundcolor=gray!25, backgroundcolor=gray!10, roundcorner=8pt]

\begin{enumerate}
\item[$\mathbf{C}_1$] the action is not forbidden 
 
\item[$\mathbf{C}_2$] the net utility or goodness of the action is
  greater than some positive amount $\gamma$;
\item[$\mathbf{C}_{3a}$] the agent performing the action intends only
  the good effects;
\item[$\mathbf{C}_{3b}$] the agent does not intend any of the bad
  effects;
\item[$\mathbf{C}_4$] the bad effects are not used as a means to
  obtain the good effects; and
\item[$\mathbf{C}_5$] if there are bad effects, the agent would rather
  the situation be different and the agent not have to perform the
  action. That is, the action is unavoidable.
\end{enumerate}
\end{mdframed}

\end{small}

Note that while \cite{nsg_sb_dde_2017} present a formalization and
corresponding implementation and ``stopwatch'' test of the first four
clauses above, there is no formalization of $\mathbf{C}_5$.  Work
presented here will enable such a formalization.  The last clause has
been rewritten below to make explicit its counterfactual nature.

\begin{small}

\begin{mdframed}[linecolor=white, frametitle=$\mathbf{C}_{5}$ Broken Up , frametitlebackgroundcolor=gray!25, backgroundcolor=gray!10, roundcorner=8pt]

\begin{enumerate}

\item[$\mathbf{C}_{5a}$] The agent desires that the current situation be
  different. 

\item[$\mathbf{C}_{5b}$] The agent believes that if the agent itself
  were in a different situation, the agent would not perform the
  action $\alpha$.

\end{enumerate}
\end{mdframed}
\end{small}

Separately, \cite{migliore2014counterfactual} have an empirical study
in which they elicit subjects to produce counterfactual answers to
questions in a mix of situations with and without moral content.
Their answers have the form of $\mathbf{C}_{5a}$ and
$\mathbf{C}_{5b}$.  Their study shows with statistical significance
that humans spent more time responding to situations that had moral
content.  This suggests the presence of non-trivial counterfactual
reasoning in morally-charged situations.

Counterfactual reasoning also plays an important role in the
intelligence community in \textbf{counterfactual forecasting}
\cite{iarpa_2017_focus}.  In counterfacutal forecasting, analysts try
to forecast what would have happened if the situation in the past was
different than what we know, and as \citeauthor{iarpa_2017_focus}
states there is a need for formal/automated tools for counterfactual
reasoning.



\section{Prior Art}

Most formal modeling of counterfactuals has been in work on
\textbf{subjunctive conditionals}.  While there are varying
definitions of what a subjunctive conditional is, the hallmark of such
conditionals is that the antecedent, while not necessarily contrary to
established facts (as is the case in counterfactuals), does speak of
what \emph{could} hold even if it presently doesn't; and then the
consequent expresses that which would (at least supposedly) hold were
this antecedent to obtain.\footnote{E.g., \english{If you were to
    practice every day, your serve would be reliable} is a subjunctive
  conditional.  It might not be the the case that you're not already
  practicing hard.  However, \english{If you had practiced hard, your
    serve would have been reliable} is a counterfactual (because, as
  it's said in the literature, the antecedent is ``contrary to
  fact'').} Hence, to ease exposition herein, we simply note that (i)
subjunctives are assuredly non-truth-functional conditionals, and (ii)
we can take subjunctive conditionals to be a superclass of
counterfactual conditionals.
A lively overview of formal systems for modeling subjunctive
conditionals can be found in \cite{nute_conditional_logic}.  Roughly,
prior work can be divided into \emph{cotenability theories} versus
\emph{possible-worlds} theories.  In \emph{cotenability theories}, a
subjunctive $\phi > \psi$ holds \emph{iff} $(\mathbf{C} + \phi)
\rightarrow \psi$ holds. Here $C$ is taken to be a set of laws
(logical/physical/legal) cotenable with $\phi$.  One major issue with
many theories of cotenability is that they at least have the
appearance of circularly defining cotenability in terms of
cotenability.  In \emph{possible-worlds} theories, semantics of
subjunctive conditionals are defined in terms of possible
worlds.\footnote{E.g., \cite{lewis_counterfactuals} famously aligns
  each possible world with an order of relative similarity among
  worlds, and is thereby able to capture in clever fashion the idea
  that a counterfactual $\phi > \psi$ holds just in case the possible
  world satisfying $\phi$ that is the most similar to the actual world
  also satisfies $\psi$.  While as is plain we are not fans of
  possible-worlds semantics, those attracted to such an approach to
  counterfactuals would do well in our opinion to survey
  \cite{counterfactuals_fritz_goodman}.}
While conceptually attractive to a degree, such approaches are
problematic.  For example, many possible-worlds accounts are
vulnerable to proofs that certain conceptions of possible worlds are
provably inconsistent (e.g.\ see
\cite{are_there_set-theoretic_worlds}).  For detailed argumentation
against possible-world semantics for counterfactual conditionals, see
\cite{ellis1977objection}.



Relevance logics strive to fix issues such as explosion and
non-relevance of antecedents and consequents in material conditionals;
see \cite{sep-logic-relevance2017} for a wide-ranging overview.  The
main concern in relevance logics, as the name implies, is to ensure
that there is some amount of relevance between an antecedent and a
consequent of a conditional, and between the assumptions in a proof
and its conclusions.  Our model does not reflect this concern, as
common notions of relevance such as variable/expression sharing become
muddled when the system includes equality, and become even more
muddled when intensionality is added.  Most systems of relevance logic
are primarily possible-worlds-based and share some of the same
concerns we have discussed above.  For example,
\cite{mares1995relevant,mares2004relevant} discuss relevance logics
that can handle counterfactual conditionals but are all based on
possible-worlds semantics, and the formulae are only extensional in
nature.\footnote{By \textbf{extensional} logics, we refer broadly to
  non-modal logics such as propositional logic, first-order logic,
  second-order logic etc. By \textbf{intensional} logics, we refer to
  modal logics. Note that this is different from intentions which can
  be represented by intensional operators, just as knowledge, belief,
  desires etc can be represented by intensional operators. See
  \cite{zalta1988intensional} for an overview. } Work in
\cite{pereira2016counterfactuals} falls under extensional systems,
and as we explain below, is not sufficient for our modeling
requirements.

Differently, recent work in natural language processing by
\cite{son2017recognizing} focuses on detecting counterfactuals in
social-media updates.  Due to the low base rate of counterfactual
statements, they use a combined rule-based and statistical method for
detecting counterfactuals.  Their work is on detecting (and not
evaluating, analyzing further, or reasoning over) counterfactual
conditionals and other counterfactual statements.

\section{ Needed Expressivity}

Our modeling goals require a formal system $\mathcal{F}$ of adequate
expressivity to be used in moral and other theory-of-mind reasoning
tasks. $\mathcal{F}$ should be free of any consistency or soundness
issues.  In particular, $\mathcal{F}$ needs to avoid inconsistencies
such as the one demonstrated below, modified from
\cite{nsg_sb_dde_2017}.  In the inference chain below, we have an
agent $a$ who knows that the murderer is the person who owns the gun.
Agent $a$ does not know that agent $\mathit{m}$ is the murderer, but
it's true that $\mathit{m}$ is the owner of the gun.  If the knowledge
operator $\mathbf{K}$ is a simple first-order logic predicate, we get
the proof shown below, which produces a contradiction from sound
premises.  
\begin{scriptsize}
\begin{mdframed}[frametitle=Modeling Knowlege (or any Intension) in First-order
  Logic ,
  frametitlebackgroundcolor=gray!25,linecolor=white,backgroundcolor=gray!10]
\begin{scriptsize}
\vspace{-0.05in}
\begin{equation*}
\begin{aligned}
&\fbox{1}\ \ \mathbf{K}\left(a,\
  \mathsf{Murderer}\left(\mathit{owner}\left(\mathit{gun}\right)\right)\right) \mbox{
  {\color{gray}; given}} \\
&\fbox{2}\ \ \lnot \mathbf{K}\left(a,\mathsf{Murderer}\left(\mathit{m}\right)\right) \mbox{
  {\color{gray}; given}}\\
&\fbox{3}\ \ \mathit{m} = \mathit{owner}\left(\mathit{gun}\right)  \mbox{
  {\color{gray}; given}}\\
&\fbox{4}\ \ \mathbf{K}\left(a,\mathsf{Murderer}\left(\mathit{m}\right)\right)  \mbox{
  {\color{gray}; first-order inference from \fbox{3} and \fbox{1}}}\\
& \fbox{5}\ \ \mathbf{\bot}  \mbox{
  {\color{gray}; first-order inference from \fbox{4} and \fbox{2}}}
\end{aligned}
\end{equation*}
\vspace{-0.05in}
\end{scriptsize}
\end{mdframed}
\end{scriptsize}
Even more robust representation schemes can still result in such
inconsistencies, or at least unsoundness, if the scheme is extensional
in nature \cite{selmer_naveen_metaphil_web_intelligence}.  Issues such
as this arise due to uniform handling of terms that refer to the same
object in all contexts.  This is prevented if the formal system
$\mathcal{F}$ is a quantified modal logic (and other sufficiently
expressive \textbf{intensional} systems).  We present one such
quantified modal logic below.


\section{Background: Formal System}

In this section, we present the formal system in which we model
counterfactual conditionals. The formal system we use is
\textbf{deontic cognitive event calculus} (\DCEC). Arkoudas and
Bringsjord \shortcite{ArkoudasAndBringsjord2008Pricai} introduced, for
their modeling of the \emph{false-belief task}, the general family
of \textbf{cognitive event calculi} to which \DCEC\ belongs.  \DCEC\
has been used to formalize and automate highly intensional moral
reasoning processes and principles, such as \textit{akrasia} (giving
in to temptation that violates moral principles)
\cite{akratic_robots_ieee_n}. and the doctrine of double effect
described above.\footnote{ The description of \DCEC\ here is mostly a subset of
the discussion in \cite{nsg_sb_dde_2017} relevant for us.}

Briefly, \DCEC\ is a sorted (i.e.\ typed) quantified modal logic (also
known as sorted first-order modal logic).  The calculus has a
well-defined syntax and proof calculus; outlined below.  The
proof calculus is based on natural deduction
\cite{gentzen_investigations_into_logical_deduction}, commonly used by
practicing mathematicians and logicians, as well as to teach logic;
the proof calculus includes all the standard introduction and
elimination rules for first-order logic, as well as inference schemata
for the modal operators and related structures.

\subsection{Syntax}
\label{subsect:syntax}

\DCEC\ is a sorted calculus.  A sorted system can be regarded
analogous to a typed single-inheritance programming language.  We show
below some of the important sorts used in \DCEC.  Among these, the
\type{Agent}, \type{Action}, and \type{ActionType} sorts are not
native to the event calculus.

The syntax can be thought of as having two components: a first-order
core and a modal system that builds upon this first-order core.  The
figures below show the syntax and inference schemata of \DCEC.  The
syntax is quantified modal logic. The first-order core of \DCEC\ is
the \emph{event calculus} \cite{mueller_commonsense_reasoning}.
Commonly used function and relation symbols of the event calculus are
included.  Other calculi (e.g.\ the \emph{situation calculus}) for
modeling commonsense and physical reasoning can be easly switched out
in-place of the event calculus.



The modal operators present in the calculus include the standard
operators for knowledge $\knows$, belief $\believes$, desire
$\desires$, intention $\intends$, etc.  The general format of an
intensional operator is $\knows\left(a, t, \phi\right)$, which says
that agent $a$ knows at time $t$ the proposition $\phi$.  Here $\phi$
can in turn be any arbitrary formula. Also,
note the following modal operators: $\mathbf{P}$ for perceiving a
state, 
$\mathbf{C}$ for common knowledge, $\mathbf{S}$ for agent-to-agent
communication and public announcements, $\mathbf{B}$ for belief,
$\mathbf{D}$ for desire, $\mathbf{I}$ for intention, and finally and
crucially, a dyadic deontic operator $\mathbf{O}$ that states when an
action is obligatory or forbidden for agents. It should be noted that
\DCEC\ is one specimen in a \emph{family} of easily extensible
cognitive calculi.
 
The calculus also includes a dyadic (arity = 2) deontic operator
$\ought$. It is well known that the unary ought in standard deontic
logic lead to contradictions.  Our dyadic version of the operator
blocks the standard list of such contradictions, and
beyond.\footnote{A overview of this list is given lucidly in
  \cite{sep_deontic_logic}.}

 \begin{scriptsize}
\begin{mdframed}[linecolor=white, frametitle=Syntax,
  frametitlebackgroundcolor=gray!25, backgroundcolor=gray!10,
  roundcorner=8pt]
\vspace{-0.1in}
\begin{equation*}
 \begin{aligned}
    \mathit{S} &::= 
    \begin{aligned}
      & \Agent \sep \ActionType \sep \Action \sqsubseteq
      \Event \sep \Moment  \sep \Fluent \\
    \end{aligned} 
    \\ 
    \mathit{f} &::= \left\{
    \begin{aligned}
      & \action: \Agent \times \ActionType \rightarrow \Action \\
      &  \initially: \Fluent \rightarrow \Boolean\\
      &  \holds: \Fluent \times \Moment \rightarrow \Boolean \\
      & \happens: \Event \times \Moment \rightarrow \Boolean \\
      & \clipped: \Moment \times \Fluent \times \Moment \rightarrow \Boolean \\
      & \initiates: \Event \times \Fluent \times \Moment \rightarrow \Boolean\\
      & \terminates: \Event \times \Fluent \times \Moment \rightarrow \Boolean \\
      & \prior: \Moment \times \Moment \rightarrow \Boolean\\
    \end{aligned}\right.\\
        \mathit{t} &::=
    \begin{aligned}
      \mathit{x : S} \sep \mathit{c : S} \sep f(t_1,\ldots,t_n)
    \end{aligned}
    \\ 
    \mathit{\phi}&::= \left\{ 
    \begin{aligned}
     & q:\Boolean \sep  \neg \phi \sep \phi \land \psi \sep \phi \lor
     \psi \sep \forall x: \phi(x) \sep \\\
 &\perceives (a,t,\phi)  \sep \knows(a,t,\phi) \sep     \\ 
& \common(t,\phi) \sep
 \says(a,b,t,\phi) 
     \sep \says(a,t,\phi) \sep  \believes(a,t,\phi) \\
& \desires(a,t,\phi)  \sep \intends(a,t,\phi) \\ & \ought(a,t,\phi,(\lnot)\happens(action(a^\ast,\alpha),t'))
      \end{aligned}\right.
  \end{aligned}
\end{equation*}
\end{mdframed}
\end{scriptsize}

\subsection{Inference Schemata}

The figure below shows a fragment of the inference schemata for \DCEC.
First-order natural deduction introduction and elimination rules are
not shown. Inference schemata $R_\mathbf{K}$ and $R_\mathbf{B}$ let us
model idealized systems that have their knowledge and beliefs closed
under the \DCEC\ proof theory.  While humans are not dedcutively
closed, these two rules lets us model more closely how more deliberate
agents such as organizations, nations and more strategic actors
reason. (Some dialects of cognitive calculi restrict the number of
iterations on intensional
operators.) 
$R_4$ states that knowledge of a proposition implies that the
proposition holds $R_{13}$ ties intentions directly to perceptions
(This model does not take into account agents that could fail to carry
out their intentions).  $R_{14}$ dictates how obligations get
translated into known intentions.

\begin{scriptsize}

\begin{mdframed}[linecolor=white, frametitle=Inference Schemata
  (Fragment), nobreak=true, frametitlebackgroundcolor=gray!25, backgroundcolor=gray!10, roundcorner=8pt]
\begin{equation*}
\begin{aligned}
  &\hspace{40pt} \infer[{[R_{\knows}]}]{\knows(a,t_2,\phi)}{\knows(a,t_1,\Gamma), \ 
    \ \Gamma\vdash\phi, \ \ t_1 \leq t_2}  \\ 
& \hspace{40pt} \infer[{[R_{\believes}]}]{\believes(a,t_2,\phi)}{\believes(a,t_1,\Gamma), \ 
    \ \Gamma\vdash\phi, \ \ t_1 \leq t_2} \\
& \hspace{20pt} \infer[{[R_4]}]{\phi}{\knows(a,t,\phi)}
\hspace{18pt}\infer[{[R_{13}]}]{\perceives(a,t', \psi)}{t<t', \ \ \intends(a,t,\psi)}\\
&\infer[{[R_{14}]}]{\knows(a,t,\intends(a,t,\chi))}{\begin{aligned}\ \ \ \ \believes(a,t,\phi)
 & \ \ \
 \believes(a,t,\ought(a,t,\phi, \chi)) \ \ \ \ought(a,t,\phi,
 \chi)\end{aligned}}
\end{aligned}
\end{equation*}
\end{mdframed}
\end{scriptsize}

\subsection{Semantics}

The semantics for the first-order fragment is the standard first-order
semantics. The truth-functional connectives
$\land, \lor, \rightarrow, \lnot$ and quantifiers $\forall, \exists$
for pure first-order formulae all have the standard first-order
semantics. The semantics of the modal operators differs from what is
available in the so-called Belief-Desire-Intention (BDI) logics
{\cite{bdi_krr_1999}} in many important ways.  For example, \DCEC\
explicitly rejects possible-worlds semantics and model-based
reasoning, instead opting for a \textit{proof-theoretic} semantics and
the associated type of reasoning commonly referred to as
\textit{natural deduction}
\cite{gentzen_investigations_into_logical_deduction,proof-theoretic_semantics_for_nat_lang}.
Briefly, in this approach, meanings of modal operators are defined via
arbitrary computations over proofs, as we will see for the
counterfactual conditional below.

\subsection{Reasoner (Theorem Prover)} Reasoning is performed through
the first-order modal logic theorem prover, \textsf{ShadowProver} \cite{nsg_sb_dde_2017}.
While we use the prover in our simulations, describing the prover in
more detail is out of scope for the present paper.\footnote{The underlying first-order prover is
  SNARK \cite{snark.94.cade}.}
%
%


\section{The Formal Problem}

At a time point $t$, we are given a set of sentences $\Gamma$
describing what the system (or agent at hand) is working with. This
set of sentences can be taken to describe the current state of the
world at $t$; state of the world before $t$, up-to to some horizon $h$
in the past; and also desires and beliefs about the future.  We are
given a counterfactual conditional $\phi \hookrightarrow \psi$ with
possibly (but not always) $\Gamma \vdash \lnot \phi$.  We require that
our formal model provide us with the following:

\begin{small}

\begin{mdframed}[linecolor=white, frametitle= Required Conditions, frametitlebackgroundcolor=gray!25, backgroundcolor=gray!10, roundcorner=8pt]
\begin{enumerate}

\item Given a set of formulae $\Gamma$, and a sentence of the form
  $\counterfac{\phi}{\psi}$, we should be able to tell whether or not
  $\Gamma\vdash \counterfac{\phi}{\psi}$.

\item Given a set of formulae $\Gamma$, and a sentence of the form
  $\mathbf{\Theta}(a, t, \counterfac{\phi}{\psi})$, we should be able to tell whether or not
  $\Gamma\vdash \mathbf{\Theta}(a, t, \counterfac{\phi}{\psi})$, here
  $\mathbf{\Theta}$ is either $\mathbf{B}$, $\mathbf{K}$ or $\mathbf{D}$.
\end{enumerate} 
\end{mdframed}
\end{small}

\subsection*{On Using the Material Conditional} If we are
considering a simple material conditional $\phi \rightarrow \psi$,
then if $\Gamma\vdash\lnot \phi$, then trivially
$\Gamma\vdash \phi \rightarrow \psi$, if the proof calculus subsumes
standard propositional logic $\vdash_{\mathsf{Prop}}$, as is is the case with logics used in
moral reasoning. Another issue is that whether
$\Gamma\vdash (\counterfac{\phi}{\psi})$ holds is not simply a function of
whether $\Gamma\vdash\phi$ holds and $\Gamma\vdash\psi$ holds, that is
$\hookrightarrow$ is not a truth-functional connective, unlike the
material conditional $\rightarrow$.

\section{Modeling Counterfactual Conditionals}

The general intuition is that given $\counterfac{\phi}{\psi}$, one has to
drop some of the formulae in $\Gamma$ arriving at $\Gamma'$ and then
if $(\Gamma + \phi) \vdash \psi$, we can conclude that
$\counterfac{\phi}{\psi}$. 
More precisely, $\counterfac{\phi}{\psi}$ can be proven from $\Gamma$ \emph{iff} there is
a subset $\Gamma'$of $\Gamma$ consistent with $\phi$ such that $(\Gamma' + \phi) \vdash \psi$.
Since we are using a natural deduction
proof calculus, we need to specify introduction and elimination rules
for $\hookrightarrow$. Let $\mathsf{Cons}[\Phi]$ denote that a set of
formulae $\Phi$ is consistent.

\subsection{Extensional Context}

\begin{small}

\begin{mdframed}[linecolor=white, frametitle=  $R_{\mathsf{cf}_1}\  \ \hookrightarrow$
  Introduction, frametitlebackgroundcolor=gray!25,
  backgroundcolor=gray!10, roundcorner=8pt]
\vspace{-0.1in}
\begin{equation*}
\begin{aligned}
\Big(\Gamma \vdash \counterfac{\phi}{\psi}\Big) \
\mathit{\Leftrightarrow}  \lnot \ConsF{\phi} \lor  \exists \Gamma' \left\{
\begin{aligned} \ 
   [\mathbf{I}_1] & \ \Gamma' \subseteq\Gamma \\
   [\mathbf{I}_2] & \ \mathsf{Con}[\Gamma' + \phi]\\
[\mathbf{I}_3] & \ (\Gamma' + \phi)\vdash \psi\\
\end{aligned}\right.
\end{aligned}
\end{equation*}
\end{mdframed}
\end{small}
The elimination rule for $\hookrightarrow$ is much simpler and
resembles the rule for the material conditional $\rightarrow$.
\begin{small}
\begin{mdframed}[linecolor=white, frametitle= $R_{\mathsf{cf}_2}\  \ \hookrightarrow$
  Elimination, frametitlebackgroundcolor=gray!25,
  backgroundcolor=gray!10, roundcorner=8pt]

\begin{equation*}
\begin{aligned}
\Gamma + \{\phi, \counterfac{\phi}{\psi}\} \vdash \psi
\end{aligned}
\end{equation*}
\end{mdframed}
\end{small}

\subsection{Intensional Context}

The inference schema given below apply to the presence of the
counterfactual in any arbitrary context $\Upsilon$.  The context for a
formula is defined as below. Let $\langle\rangle$ denote the empty list
and let $\oplus$ denote list concatentation. 

\begin{small}
\begin{mdframed}[linecolor=white, frametitle= $\hookrightarrow$
  Definition of $\Upsilon$, frametitlebackgroundcolor=gray!25,
  backgroundcolor=gray!10, roundcorner=8pt]
\begin{equation*}
\begin{aligned}
\Upsilon[\phi] =  \left\{ 
\begin{aligned} 
& \langle\rangle \mbox{ if the top connective in $\phi$ is not modal}\\
& \langle\knows, a, t\rangle,  \oplus\Upsilon[\psi],  \mbox{ if $\phi$
  }  \equiv \knows(a, t,\psi)\\
& \langle\believes, a, t\rangle,  \oplus\Upsilon[\psi],  \mbox{ if $\phi$
  }  \equiv \believes(a, t,\psi)\\
& \langle\desires, a, t\rangle,  \oplus\Upsilon[\psi],  \mbox{ if $\phi$
  }  \equiv \desires(a, t,\psi)\\
\end{aligned}\right.
\end{aligned}
\end{equation*}
\end{mdframed}
\end{small}
For example, $\Upsilon[\believes(a,t_1,\Knows(b,t_2, P))]= \langle
\believes, a,t_1, \knows, b, t_2\rangle $
With the context defined as above, inference schemata for 
$\hookrightarrow$ occurring within a modal context is given below.

We also use the following shorthands in the two rules given below:

\begin{small}
\begin{equation*}
\begin{aligned}
&(i) \Upsilon[\phi]\mbox{ denotes that formula $\phi$ has context } \Upsilon.\\
&(ii) \Upsilon[\Gamma] \mbox{ denotes the subset of $\Gamma$ with
  context } \Upsilon.
\end{aligned}
\end{equation*}
\end{small}
\begin{small}

\begin{mdframed}[linecolor=white, frametitle= $R_{\mathsf{cf}_3}\  \hookrightarrow$
  Introduction, frametitlebackgroundcolor=gray!25,
  backgroundcolor=gray!10, roundcorner=8pt]
\begin{equation*}
\begin{aligned}
\!\!\!\!\big(\Gamma \vdash \Upsilon\left[\counterfac{\phi}{\psi}\right]\big) \
\mathit{\Leftrightarrow}\  \lnot \ConsF{\phi} \lor  \exists \Gamma' \left\{
\begin{aligned} \ 
   [\mathbf{I}_1] & \ \Gamma' \subseteq \Upsilon[\Gamma]\\
   [\mathbf{I}_2] & \ \mathsf{Con}[\Gamma' + \phi]\\
[\mathbf{I}_3] & \ (\Gamma' + \phi)\vdash \psi\\
\end{aligned}\right.
\end{aligned}
\end{equation*}
\end{mdframed}
\end{small}
The elimination rule for $\hookrightarrow$ under any arbitrary context
is given below.

\begin{small}

\begin{mdframed}[linecolor=white, frametitle= $R_{\mathsf{cf}_4}\  \hookrightarrow$
  Elimination, frametitlebackgroundcolor=gray!25,
  backgroundcolor=gray!10, roundcorner=8pt]
\begin{equation*}
\begin{aligned}
\Gamma + \{\Upsilon[\phi], \Upsilon[\counterfac{\phi}{\psi}]\} \vdash \Upsilon[\psi]
\end{aligned}
\end{equation*}
\end{mdframed}
\end{small}
For a simple example of the above rules, see the experiments sections below.

\subsection{A Note on Implementing the Introduction Rules} There are
two possible dynamic programming algorithms. In the worst case, both
the algorithms amount to searching over all possible subsets of a
given $\Gamma$. In the first algorithm, we start from smaller subsets
and compute whether $(\Gamma'+\phi)\vdash\psi$ and
$\Cons{\Gamma'+\phi}$. For any larger sets $\Gamma''$,
$(\Gamma'' +\phi)\vdash\psi$ need not be computed. In the second
algorithm, we start with larger subsets and compute
$(\Gamma''+\phi)\vdash\psi$ and $\Cons{\Gamma''+\phi}$. For any
smaller sets $\Gamma'$, $\Cons{\Gamma'+\phi}$ need not be computed. If
$\vdash$ is first-order and above, the second algorithm is
preferable. If $\vdash$ encompasses first-order logic, then computing
$\Cons{\Phi}$ for any set is hard in the general case. In our
implementation, we approximate $\Cons{\Phi}$ by running querying a
prover for $\Phi\vdash\bot$ up to a time limit $\delta$. As $\delta$
increases, the approximation becomes better.


\subsection{Properties of the System}
\newcounter{propcounter}
\newcounter{thmcounter}

Now we canvass some meta-theorems about the system.  All the four
inference schemata given above, when added to a monotonic proof
theory, preserve monotonicity.

\begin{small}
\stepcounter{thmcounter}
\begin{mdframed}[linecolor=white, frametitle=  Theorem \thethmcounter:
  Monotonicity Preservation, frametitlebackgroundcolor=gray!25,
  backgroundcolor=gray!10, nobreak=true ,roundcorner=8pt]
\begin{equation*}
\begin{aligned}
&\mbox{If } \vdash \mbox{ is monotonic, then } \vdash \mbox{
  augmented with }\\ &R_{\mathsf{cf}_1}, R_{\mathsf{cf}_2}, R_{\mathsf{cf}_3}, \mbox{ and }R_{\mathsf{cf}_4}
\mbox{ is still monotonic}. 
\end{aligned}
\end{equation*}
\end{mdframed}
\end{small}

\noindent{\textbf{Proof Sketch}}: Notice that the right-hand side of
the condition for the introduction rules stay satisfied if we replace
$\Gamma$ with a superset $\Phi$.  The elimination rules hold
regardless of $\Gamma$.  $\blacksquare$

We assume that $\vdash$ is monotonic.  For monotonic systems, for any
$\Gamma'\subseteq \Gamma$, $\Cons{\Gamma} \Rightarrow \Cons{\Gamma'}$.
We show that some desirable properties that hold for other systems in
the literature hold for the system presented here.

\begin{small}
\stepcounter{propcounter}
\begin{mdframed}[linecolor=white, frametitle= Property \thepropcounter
  \textnormal{; $\mathsf{ID}$  in \cite{nute_conditional_logic}}, frametitlebackgroundcolor=gray!25,
  backgroundcolor=gray!10, nobreak=true ,roundcorner=8pt]
\begin{equation*}
\begin{aligned}
\{ \} \vdash \counterfac{\phi}{\phi}
\end{aligned}
\end{equation*}
\end{mdframed}
\end{small}

\noindent{\textbf{Proof}}: If $\lnot\ConsF{\phi}$, we are done.
Otherwise, since $\{\phi\}\vdash \phi$, we can see that \Ia, \Ib, and
\Ic\ hold. $\blacksquare$

\begin{small}
\stepcounter{propcounter}
\begin{mdframed}[linecolor=white, frametitle= Property \thepropcounter
  \textnormal{; $\mathsf{R2}$  in \cite{Pollock1976-POLSR}}, frametitlebackgroundcolor=gray!25,
  backgroundcolor=gray!10, nobreak=true ,roundcorner=8pt]
\begin{equation*}
\begin{aligned}
\big(\{\} \vdash \matcond{\phi}{\psi}\big) \Rightarrow \big(\{\} \vdash \counterfac{\phi}{\psi}\big)
\end{aligned}
\end{equation*}
\end{mdframed}
\end{small}

\noindent{\textbf{Proof}}: Either $\lnot\ConsF{\phi}$ or
$\ConsF{\phi}$.  If the former holds, we are done.  If the latter
holds, then take $\Gamma' = \{\}$ and we have $\Cons{\Gamma'+\phi}$
and $\Gamma' + \phi \vdash \psi$, satisfying all three conditions \Ia,
\Ib, and \Ic.  $\therefore$ letting us introduce
$\counterfac{\phi}{\psi}$. $\blacksquare$


\begin{small}
\stepcounter{propcounter}
\begin{mdframed}[linecolor=white, frametitle= Property \thepropcounter
   \textnormal{; $\mathsf{R3}$  in \cite{Pollock1976-POLSR}}, frametitlebackgroundcolor=gray!25,
  backgroundcolor=gray!10, nobreak=true ,roundcorner=8pt]
\begin{equation*}
\begin{aligned}
\Big(\{\} \vdash \matcond{\phi}{\psi}\Big) \Rightarrow \Big(\{\} \vdash \matcond{(\counterfac{\chi}{\phi})}{(\counterfac{\chi}{\psi})}\Big)
\end{aligned}
\end{equation*}
\end{mdframed}
\end{small}

\noindent{\textbf{Proof}}: Assume $\{\} \vdash \matcond{\phi}{\psi}$.
We need to prove $\{\counterfac{\chi}{\phi}\}\vdash
(\counterfac{\chi}{\psi})$.  If
$\lnot\ConsF{\counterfac{\chi}{\phi}}$, we succeed.  Otherwise, take
$\Gamma'$ to be $\{\counterfac{\chi}{\phi}\}$.  By the elimination
rule we have $\{\counterfac{\chi}{\phi}\} + \chi \vdash \phi$, and
using $\{\}\vdash \matcond{\phi}{\psi}$, we have
$\{\counterfac{\chi}{\phi}\} + \chi \vdash \psi$, thus satisfying our
conditions \Ia, \Ib, and \Ic\ to introduce
$\counterfac{\chi}{\psi}$. $\blacksquare$


\begin{small}
\stepcounter{propcounter}
\begin{mdframed}[linecolor=white, frametitle= Property \thepropcounter
   \textnormal{; $\mathsf{A6}$  in \cite{Pollock1976-POLSR}, $\mathsf{MP}$  in \cite{nute_conditional_logic}, }, frametitlebackgroundcolor=gray!25,
  backgroundcolor=gray!10, nobreak=true ,roundcorner=8pt]
\begin{equation*}
\begin{aligned}
\mbox{If }\Gamma \vdash \counterfac{\phi}{\psi} \mbox{, then } \Gamma
\vdash {\phi} \rightarrow {\psi} 
\end{aligned}
\end{equation*}
\end{mdframed}
\end{small}

\noindent{\textbf{Proof}}: $\Gamma \vdash (\counterfac{\phi}{\psi})$,
therefore $\Gamma + \{\phi\} \vdash {\psi}$ using \elim{\hookrightarrow}.
Using \intro{\rightarrow} we have
$\Gamma \vdash {\phi}\rightarrow{\psi}$. $\blacksquare$


\begin{small}
\stepcounter{propcounter}
\begin{mdframed}[linecolor=white, frametitle= Property \thepropcounter
  \textnormal{; $\mathsf{MOD}$  in \cite{nute_conditional_logic}}, frametitlebackgroundcolor=gray!25,
  backgroundcolor=gray!10, nobreak=true ,roundcorner=8pt]
\begin{equation*}
\begin{aligned}
\mbox{If }\Gamma \vdash \counterfac{(\lnot \psi)}{\psi} \mbox{, then } \Gamma
\vdash {\phi} \rightarrow {\psi} 
\end{aligned}
\end{equation*}
\end{mdframed}
\end{small}

\noindent{\textbf{Proof}}: Given $\Gamma \vdash \counterfac{(\lnot
  \psi)}{\psi}$, therefore using \elim{\hookrightarrow} $\Gamma +
\lnot \psi\vdash \psi$, and using \elim{\lnot} we have $\Gamma \vdash
\psi$, giving us $\Gamma \vdash \phi \rightarrow \psi$. $\blacksquare$


\begin{small}
\stepcounter{propcounter}
\begin{mdframed}[linecolor=white, frametitle= Property \thepropcounter
 , frametitlebackgroundcolor=gray!25,
  backgroundcolor=gray!10, nobreak=true ,roundcorner=8pt]
\begin{equation*}
\begin{aligned}
&\mbox{If }  \Gamma\vdash (\counterfac{\phi}{\psi}) \land
(\counterfac{\psi}{\phi})
\\ &\mbox{ then } \Gamma\vdash 
({\phi}\rightarrow {\chi})\ \Leftrightarrow  \Gamma \vdash
({\psi}\rightarrow {\chi})
\end{aligned}
\end{equation*}
\end{mdframed}
\end{small}

\noindent{\textbf{Proof}}: We prove the left-to-right direction of the
biconditional
$$\Gamma\vdash (\matcond{\phi}{\chi})\ \Rightarrow \Gamma \vdash
(\matcond{\psi}{\chi}) $$ 
Assume $\Gamma\vdash (\matcond{\phi}{\chi})$.  We need to prove
$\Gamma\vdash (\matcond{\psi}{\chi})$.  Since we are given $\Gamma
\vdash \counterfac{\psi}{\phi},$ either $\lnot \ConsF{\psi}$ or there
is a $\Gamma'\subseteq\Gamma$ such that $\Cons{\Gamma' + \psi}$ and
$\Gamma' + \psi\ \vdash \phi$.  If the former holds, we are finished.
If the latter holds, we then have $\Gamma + \psi \vdash \phi$ since
$\vdash$ is monotonic.  Using the given $\Gamma\vdash
(\matcond{\phi}{\chi})$ and the $\rightarrow-$elimination rule, we
obtain $\Gamma + \psi \vdash \chi$.  Therefore, $\Gamma\vdash
\matcond{\psi}{\chi}$. $\blacksquare$

\begin{small}
\stepcounter{propcounter}
\begin{mdframed}[linecolor=white, frametitle= Property \thepropcounter
  \textnormal{; $\mathsf{CSO}$  in \cite{nute_conditional_logic}}, frametitlebackgroundcolor=gray!25,
  backgroundcolor=gray!10, nobreak=true ,roundcorner=8pt]
\begin{equation*}
\begin{aligned}
&\mbox{If } \begin{aligned} & \EmptySet \vdash (\counterfac{\phi}{\psi}) \land
(\counterfac{\psi}{\phi}) \end{aligned}
\\ &\mbox{ then } \Gamma\vdash 
(\counterfac{\phi}{\chi})\ \Leftrightarrow  \Gamma \vdash (\counterfac{\psi}{\chi})
\end{aligned}
\end{equation*}
\end{mdframed}
\end{small}

\noindent{\textbf{Proof}}: We establish the left-to-right direction of
the biconditional
$$\Gamma\vdash (\counterfac{\phi}{\chi})\ \Rightarrow \Gamma \vdash
(\counterfac{\psi}{\chi})$$
Assuming $\EmptySet\vdash (\counterfac{\phi}{\psi}) \land
(\counterfac{\psi}{\phi})$ and using \textbf{Property 4} and
$\vdash_\mathsf{Prop}$, we have $\EmptySet\vdash
(\matcond{\phi}{\psi}) \land (\matcond{\psi}{\phi})$; and using
$\vdash_{\mathsf{Prop}}$ again we obtain $\EmptySet\vdash\phi
\leftrightarrow \psi$.  Assume $\Gamma\vdash
(\counterfac{\phi}{\chi})$.  We need to prove $\Gamma\vdash
(\counterfac{\psi}{\chi})$.  Since we assumed $\Gamma \vdash
\counterfac{\phi}{\chi},$ either $\lnot \ConsF{\phi}$ or there is a
$\Gamma'\subseteq\Gamma$ such that $\Cons{\Gamma' + \phi}$ and
$\Gamma' + \phi\ \vdash \chi$.  If the former holds we are done
because $\lnot\ConsF{\phi} \Rightarrow \lnot\ConsF{\psi} $.  If the
latter holds, we have $\Cons{\Gamma' + \psi}$ and $\Gamma' +
\psi\vdash \chi$ as $\{\}\vdash \phi \leftrightarrow \psi$.
$\blacksquare$


\begin{small}
\stepcounter{propcounter}
\begin{mdframed}[linecolor=white, frametitle= Property \thepropcounter, frametitlebackgroundcolor=gray!25,
  backgroundcolor=gray!10, nobreak=true ,roundcorner=8pt]
\begin{equation*}
\begin{aligned}
&\mbox{If } \begin{aligned} & \EmptySet \vdash (\counterfac{\phi}{\psi}) \land
(\counterfac{\psi}{\phi}) \end{aligned}
\\ &\mbox{ then } \Gamma\vdash 
(\counterfac{\chi}{\phi})\ \Leftrightarrow  \Gamma \vdash (\counterfac{\chi}{\psi})
\end{aligned}
\end{equation*}
\end{mdframed}
\end{small}

\noindent{\textbf{Proof}}: Similar to the previous proof, we prove the
left-to-qright direction of the biconditional
$$\Gamma\vdash (\counterfac{\phi}{\chi})\ \Rightarrow \Gamma \vdash
(\counterfac{\psi}{\chi})$$
Assuming $\EmptySet\vdash (\counterfac{\phi}{\psi}) \land
(\counterfac{\psi}{\phi})$ and using \textbf{Property 4} and
$\vdash_\mathsf{Prop}$, we deduce $\EmptySet\vdash
(\matcond{\phi}{\psi}) \land (\matcond{\psi}{\phi})$; and using
$\vdash_{\mathsf{Prop}}$ again we infer $\EmptySet\vdash\phi
\leftrightarrow \psi$.  Assume $\Gamma\vdash
(\counterfac{\chi}{\phi})$.  We need to show $\Gamma\vdash
(\counterfac{\chi}{\psi})$.  Since we assumed $\Gamma \vdash
\counterfac{\chi}{\phi},$ either $\lnot \ConsF{\chi}$ or there is a
$\Gamma'\subseteq\Gamma$ such that $\Cons{\Gamma' + \chi}$ and
$\Gamma' + \chi\ \vdash \phi$.  If the former, we are done; if the
latter, we have $\Gamma' + \chi\vdash \psi$.  $\blacksquare$

The next property states that if we are given
$\counterfac{\phi}{\psi_1}$ and $\counterfac{\phi}{\psi_2}$, we can
reach $\counterfac{\phi}{\psi_1\land \psi_2}$.  For instance,
\english{If a had gone to the doctor, a would not have been sick} and
\english{If a had gone to the doctor, a would be happy now} give us
\english{If a had gone to the doctor, a would not have been sick and a
  would be happy now}.

One property which has been problematic for many previous formal
systems is \textbf{simplification of disjunctive antecedents}
(\textsf{SDA}). For instance, given \english{If a had gone to the
  doctor or if a had taken medication, a would not be sick now} should
give us \english{If a had gone to the doctor, a would not be sick now,
  and if a had taken medication, a would not be sick now.} Logics
based on possible worlds find it quite challenging to handle this
property --- yet it seems like a necessary property to secure
\cite{ellis1977objection}, though there are some disagreements on this
front \cite{loewer1976counterfactuals}.  As a middle ground we can
prove \english{If a had gone to the doctor, a would not be sick now,
  \textbf{or} if a had taken medication, a would not be sick now.}


\begin{small}
\stepcounter{propcounter}
\begin{mdframed}[linecolor=white, frametitle= Property \thepropcounter, frametitlebackgroundcolor=gray!25,
  backgroundcolor=gray!10, nobreak=true ,roundcorner=8pt]
\begin{equation*}
\begin{aligned}
\mbox{If } &\Gamma\vdash \counterfac{\left(\phi_1 \lor \phi_2\right)}{\psi}
\mbox { then } 
\\ &\Gamma\vdash (\counterfac{\phi_1}{\psi})\lor (\counterfac{\phi_1}{\psi})
\end{aligned}
\end{equation*}
\end{mdframed}
\end{small}

\noindent{\textbf{Proof}}: We need to either prove $\Gamma\vdash
\counterfac{\phi_1}{\psi}$ or $\Gamma\vdash \counterfac{\phi_1}{\psi}$
and use $\lor-$introduction to get $\Gamma\vdash
(\counterfac{\phi_1}{\psi})\lor(\counterfac{\phi_2}{\psi}).$ If
$\lnot\Cons{\phi_1 \lor \phi_2}$, we have $\lnot\Cons{\phi_1}$ and
$\lnot\Cons{\phi_2}$, we then reach our conclusion.  Otherwise, if
$\Cons{\phi_1 \lor \phi_2}$, then there is a $\Gamma_{\phi_1 \lor
  \phi_2}$ serving as $\Gamma'$.  We need to prove three clauses:
\begin{enumerate}[(1)] 

\item \Ia\ is satisfied as $\Gamma' = \Gamma_{\phi_1 \lor \phi_2}$.

\item \Ib\ is satisfied because if $\Cons{\Gamma' + \phi_1 \lor
  \phi_2}$ then at least one of $\Cons{\Gamma' + \phi_1}$ or
  $\Cons{\Gamma' + \phi_2}$; otherwise if $\Gamma'+\phi_1\vdash \bot$
  and $\Gamma'+\phi_2\vdash \bot$, giving us $\Gamma'+(\phi_1 \lor
  \phi_2)\vdash \bot$ through $\elim{\lor}$, which overthrows our
  assumption that $\Cons{\Gamma' + (\phi_1 \lor \phi_2)}.$  So either
  $\Cons{\Gamma'+\phi_1}$ or $\Cons{\Gamma'+\phi_2}$, then \Ib\ is
  satisfied for either $\counterfac{\phi_1}{\psi}$ or
  $\counterfac{\phi_2}{\psi}$.  Assume that it holds for the former.

\item \Ic\ is satisfied as $\{\phi_1\}\vdash \phi_1 \lor \phi_2$ and
  we have $\Gamma' + (\phi_1 \lor \phi_2) \vdash \psi$; therefore
  $\Gamma' + \phi_1 \vdash \psi$.  $\blacksquare$

\end{enumerate}

With stronger assumptions we can obtain a property that directly
resembles \textsf{SDA}.

\begin{small}
\stepcounter{propcounter}
\begin{mdframed}[linecolor=white, frametitle= Property \thepropcounter, frametitlebackgroundcolor=gray!25,
  backgroundcolor=gray!10, nobreak=true ,roundcorner=8pt]
\begin{equation*}
\begin{aligned}
& \mbox{If } \left(\begin{aligned} &\Gamma\vdash\counterfac{ \left(\phi_1 \lor \phi_2\right)}{\psi}\\ &\mbox{
  and } \Gamma\not\vdash\lnot \phi_1 \mbox{
  and } \Gamma\not\vdash\lnot \phi_2 \end{aligned}\right)
\\ & \mbox { then } 
\Gamma\vdash (\counterfac{\phi_1}{\psi})\land (\counterfac{\phi_1}{\psi})
\end{aligned}
\end{equation*}
\end{mdframed}
\end{small}

\noindent{\textbf{Proof}}: The proof is similar to the previous proof,
needing only minor modifications. \end{enumerate}


 \begin{small}
\stepcounter{propcounter}
\begin{mdframed}[linecolor=white, frametitle= Property \thepropcounter
 \textnormal{, $\mathsf{A4}$  in \cite{Pollock1976-POLSR}}, frametitlebackgroundcolor=gray!25,
  backgroundcolor=gray!10, nobreak=true ,roundcorner=8pt]
\begin{equation*}
\begin{aligned}
& \mbox{If } \Gamma\vdash (\counterfac{\phi}{\psi}) \mbox{ and }
\Gamma\vdash(\counterfac{\phi}{\chi})\\
&\mbox{then } \Gamma\vdash \counterfac{(\phi \land \psi)}{\chi}
\end{aligned}
\end{equation*}
\end{mdframed}
\end{small}

\noindent{\textbf{Proof}}: Since we have $\Gamma \vdash
(\counterfac{\phi}{\chi})$, either $\lnot\ConsF{\phi}$ or there is a
$\Gamma'\subseteq\Gamma$ such that $\Cons{\Gamma' +
  \phi}\vdash\chi$.  If the former holds, we have $\lnot\ConsF{ \phi
  \land \psi},$ giving us $\Gamma\vdash \counterfac{(\phi \land
  \psi)}{\chi}$.  If the latter holds, we have $\Gamma' + (\phi \land
\psi) \vdash \chi$.  Assume that $\lnot \Cons{\Gamma' + (\phi \land
  \psi)}$; then we get $\Gamma \cup \{\phi, \psi \} \vdash \bot,$
giving us $\Gamma + \phi \vdash \lnot \psi$ through
$\lnot-$introduction.  With $\rightarrow-$ introduction we get $\Gamma
\vdash \matcond{\phi}{\lnot\psi}$, but using $\mathbf{Property\ 4}$
and the given $\Gamma\vdash \counterfac{\phi}{\psi}$, we have
$\Gamma\vdash\matcond{\phi}{\psi}$.  Therefore, $\Gamma\vdash \lnot
\phi$.  This goes against $\Cons{\Gamma' + \phi}$, giving us that
$\Cons{\Gamma' + (\phi \land \psi)}.$ This satisfies \Ia, \Ib, and
\Ic\, which yields $\Gamma\vdash \counterfac{(\phi \land
  \psi)}{\chi}$. $\blacksquare$


Finally, we have the following theorem, the proof of which can be
obtained by exploiting all the properties above.

\begin{small}
\stepcounter{thmcounter}
\begin{mdframed}[linecolor=white, frametitle= Theorem \thethmcounter, frametitlebackgroundcolor=gray!25,
  backgroundcolor=gray!10, nobreak=true ,roundcorner=8pt]
  All the above properties hold if every counterfactual
  $\counterfac{\phi}{\psi}$ is replaced by
  $\Upsilon[\counterfac{\phi}{\psi}]$.
\end{mdframed}
\end{small}

Given the above, we can prove the following, using rules $R_4$ and
$R_{14}$. The following theorem gives an account of how an agent might
assert a counterfactual conditional in a moral/ethical situation.
\begin{small}
\stepcounter{thmcounter}
\begin{mdframed}[linecolor=white, frametitle= Theorem \thethmcounter, frametitlebackgroundcolor=gray!25,
  backgroundcolor=gray!10, nobreak=true ,roundcorner=8pt]

\begin{equation*}
\begin{aligned}
&\Big(\Gamma\vdash  \believes\big(a,t,\ought(a,t,\phi, \chi)\big) \land \ought(a,t,\phi,
 \chi) \Big)\\  &  \hspace{30pt} \Rightarrow   \big(\Gamma  \vdash \counterfac{\believes(a,t,\phi)}{\chi}\big)\end{aligned}
\end{equation*}
\end{mdframed}
\end{small}


\section{Experiments}
We describe two experiments that demonstrate the model in action, the
first in moral dilemmas and the second in general
situations.\footnote{Axioms for both the experiments, data and the
  reasoning system can be obtained here: \url{https://goo.gl/nDZtWX}}

\subsection{Counterfactual Conditionals in Moral Dilemmas}
We look at moral dilemmas used in \cite{nsg_sb_dde_2017}. Each
dilemma $d$ has an axiomatization $\Gamma_d$. Two such dilemmas are
axiomatized and studied in \cite{nsg_sb_dde_2017}. Assume that an
agent $a$ is facing a given dilemma $d$. The axiomatization $\Gamma_d$
includes $a$'s knowledge and beliefs about the dilemma. We show that
for each $d$, we can get $\mathbf{C}_{5a} $ and
$\mathbf{C}_{5b}$. Note that $\mathbf{C}_{5a} $ and $\mathbf{C}_{5b}$
talk about situations. Though event calculus does not directly model
situations, we can use fluents to do so. A situation is formalized as
an object of sort $\mathsf{Situation}$:
\begin{small}
$$\mathsf{Situation}\sqsubset \mathsf{Object}$$
\end{small}
We have the following additional symbols that tell us when an agent is
in a situation and what actions are possible for an agent in a
situation at a given time:
\begin{footnotesize}
\begin{equation*}
\begin{aligned}
&\ins\!: \Agent \times \Situation \rightarrow \Fluent\\
&\actionsit\!: \Agent \times \ActionType \times \Situation \times
\Moment\\
&\ \ \ \ \ \ \ \ \ \ \ \  \ \ \ \rightarrow \Boolean\\
\end{aligned}
\end{equation*}
\end{footnotesize} We have the following axiom which states that the only actions that
an agent can perform in a situation at a time are those sanctioned by the
$\Action$ predicate.
\begin{footnotesize}
\begin{equation*}
\begin{aligned}
\forall & a:\Agent, \alpha: \ActionType, t:\Moment. \\
&\left(\begin{aligned} & \hspace{32pt} \happens(\action(a,\alpha), t)
\rightarrow\\
& \exists \sigma: \Situation. \Big(\holds(\ins(a,\sigma), t) \land
\actionsit(a,\alpha, \sigma,t)\Big)
\end{aligned}\right)
\end{aligned}
\end{equation*}
\end{footnotesize}
As a warmup to $\mathbf{C}_{5b} $, we state $\mathbf{C}_{5a} $
below. Let the current agent be $\II$ and time be denoted by $t$. The
formula below states that the agent believes it is in a situation
$\sigma'$ and desires to be in a situation $\sigma'$ different from
$\sigma$ with at least one action type $\alpha$ that is not forbidden
and does not have any negative effects.\footnote{Note that this is a
  mere first formulation of a complex mental state and further
  refinements, simple and drastic, are possible and expected.} Here
$\mu:\Fluent\rightarrow \mathbb{R}$ denotes the total utility of a
fluent for all agents and all times.

\begin{small}
\begin{mdframed}[linecolor=white, frametitle= Formalization of $\mathbf{C}_{5a}$, frametitlebackgroundcolor=gray!25, backgroundcolor=gray!10, roundcorner=8pt]
\begin{footnotesize}
\vspace{-0.1in}
\begin{equation*}
\begin{aligned}
\!\!\! \!\!\! &\believes\left(\II, t, \holds\left(\ins\left(\II,
      \sigma\right)\right)\right) \land\\
\!\!\! \!\!\! &\desires \!\!\left(\II, t ,
     \begin{aligned} 
       \exists \rho
         \!\!\left[\begin{aligned} 
             &\holds\left(\ins\left(\II, \sigma\right)\right) \land \\
            &\exists \alpha\!\left[\begin{aligned} & \actionsit(\II, \alpha, \rho, t) \land\\
            & \lnot \ought(\II, t, \rho, \lnot\happens(\action(\II,\alpha),t)\\
            &\lnot \exists f \big(\left[\mu(f) < 0\right] \land
              \initiates(\alpha,f,t)\big)\\
            &\lnot \exists f \big(\left[\mu(f) > 0\right]\land \terminates(\alpha,f,t)\big)\end{aligned}\right]
         \end{aligned}\right]
      \end{aligned}\right)
\end{aligned}
\end{equation*}
\end{footnotesize}
\end{mdframed}
\end{small}

Let $\alpha_{D}$ be the $\DDE$-complaint action that the agent is
saddled with in the current situation $\sigma$.  Let
$\Theta(\sigma, t)$ be the inner statement in the desires modal
operator above. The statement below formalizes $\mathbf{C}_{5b}$ and
can be read as the agent believing that if the agent were in a
different situation with at least one action that is not forbidden and
does not have any negative effects, it would then not perform the
action $\alpha_D$ required of it by $\DDE$.

 \begin{small}
\begin{mdframed}[linecolor=white, frametitle= Formalization of
  $\mathbf{C}_{5b}$, frametitlebackgroundcolor=gray!25,
  backgroundcolor=gray!10, roundcorner=8pt]
\vspace{-0.1in}
\begin{equation*}
\begin{aligned}
&\believes\Bigg(\II, t ,
     \begin{aligned} 
        \begin{aligned} 
            \Theta(\sigma, t) \hookrightarrow  \lnot \happens\Big(\action(\II, \alpha_{D}), t+1\Big)
       \end{aligned}
      \end{aligned}\Bigg)
\end{aligned}
\end{equation*}
\end{mdframed}
\end{small}

We derive the above two conditions from: \begin{inparaenum}[(1)] \item
  axioms describing situations; and \item common knowledge
  $\common(\Phi)$ dictating that agents should only perform
  non-forbidden actions and in a non-dilemma situation will peform an
  action with only positive effects. \end{inparaenum}
$\mathbf{C}_{5a}$ takes around $780 \mathit{ms}$ and $\mathbf{C}_{5b}$
takes around $8.378 \mathit{s}$.

\subsection{Evaluation of the System}
We demonstrate the feasibility of our system by showing the model
working for a small dataset of representative problems. There are $16$
problems each with its own set of assumptions $\Gamma$. For each
problem $\Gamma$, we have a statement that is provably counterfactual,
that is $\Gamma\vdash\lnot\phi$. Using this statement, we build three
conditionals: \begin{inparaenum}[(1)] \item a counterfactual
  conditional $\counterfac{\phi}{\psi}$; \item an absurd material
  conditional $\matcond{\phi}{\bot}$; and \item an absurd
  counterfactual $\counterfac{\phi}{\bot}$. \end{inparaenum} The number of premises
  range from 2 to 15. One simple problem is shown below:
 \begin{small}
\begin{equation*}
  \begin{aligned}
&\Gamma =\left\{\begin{aligned}
 \forall x\!:\!\left(\begin{aligned} \mathit{Human}(x) \rightarrow
     \mathit{Mortal}(x) \end{aligned}\right),
 \mathit{Human}(socrates)\end{aligned}\right\}\\
 &\begin{aligned}
  &\Gamma\vdash\counterfac{\mathit{\lnot
    \mathit{Mortal}(socrates)}}{\lnot \mathit{Human}(socrates)}\\
  & \Gamma\vdash\matcond{\mathit{\lnot
    \mathit{Mortal}(socrates)}}{(P \land \lnot P)}\\
&\Gamma\not\vdash \counterfac{\mathit{\lnot
    \mathit{Mortal}(socrates)}}{(P \land \lnot P)}
\end{aligned}
\end{aligned}
\end{equation*}
\end{small}
The table below shows how much time proving the true and false
counterfactuals takes when compared with proving the absurd material
conditional for the same problem. As expected, the counterfactual
conditional takes more time than the material conditional. The absurd
counterfactual $\counterfac{\phi}{\bot}$ is merely intended as a sanity check and the large
reasoning times are expected (mainly due to timeouts), and
is not expected to be a common use case.

\begin{center}

\begin{small}
\begin{tabular}{lccc}  
\toprule
 \textbf{Formula}   & Mean (s) & Min  (s) &  Max  (s)\\
\midrule
 $\counterfac{\phi}{\psi}$ & 2.496 &  0.449   &  11.14     \\
 $\matcond{\phi}{\bot}$ & 0.169 & 0.089   & 0.341  \\
 $\counterfac{\phi}{\bot}$ & 19.7 & 1.93   & 120.67  \\
\bottomrule
\end{tabular}
\end{small}

 \end{center}



\section{Conclusion \& Future Work}
We have presented a novel formal model for counterfactual
conditionals. We have applied this model to complete a formalization
of an important ethical principle, the doctrine of double effect. We
have also provided an implementation of a reasoning system and a
dataset with counterfactual conditionals. There are three main threads
for future work. First, the reasoning algorithm is quite simple and
can be improved by looking at either hand-built or learned
heuristics. Second, we hope to leverage recent work in developing an
uncertainty system for a cognitive calculus
\cite{govindarajulu2017strength}. Finally, we hope to build a more
robust and extensive dataset of counterfactual reasoning validated by
multiple judgements from humans.

\clearpage\newpage
\bibliographystyle{aaai} 
\bibliography{main72,naveen}
\end{document}